# Deep Depth Super-Resolution : Learning Depth Super-Resolution using Deep Convolutional Neural Network

Xibin Song, Yuchao Dai, Xueying Qin

*Abstract*—Depth image super-resolution is an extremely challenging task due to the information loss in sub-sampling. Deep convolutional neural network have been widely applied to color image super-resolution. Quite surprisingly, this success has not been matched to depth super-resolution. This is mainly due to the inherent difference between color and depth images. In this paper, we bridge up the gap and extend the success of deep convolutional neural network to depth super-resolution. The proposed deep depth super-resolution method learns the mapping from a low-resolution depth image to a high resolution one in an end-to-end style. Furthermore, to better regularize the learned depth map, we propose to exploit the depth field statistics and the local correlation between depth image and color image. These priors are integrated in an energy minimization formulation, where the deep neural network learns the unary term, the depth field statistics works as global model constraint and the color-depth correlation is utilized to enforce the local structure in depth images. Extensive experiments on various depth super-resolution benchmark datasets show that our method outperforms the state-of-the-art depth image super-resolution methods with a margin.

## I. Introduction

Recently, consumer depth cameras (*e.g.* Microsoft Kinect, ASUS Xtion Pro or the Creative Senz3D camera *etc.*) and other time-of-flight (ToF) cameras have gained significant popularity due to their affordable cost and great applicability in human computer interaction [1], computer graphics and 3D modeling [2]. However, depth image outputs from these cameras suffer from natural upper limit on spatial resolution and the precision of each depth sample, thus, making depth image super-resolution (DSR) attract more and more research attentions in the community. DSR aims at obtaining a high-resolution (HR) depth image from a low-resolution (LR) depth image by inferring all the missing high frequency contents. DSR is a highly ill-posed problem as the known variables in the LR images are greatly outnumbered by the unknowns in the HR depth images.

Meanwhile, there is mounting evidences that effective features and information learned from deep convolutional

X. Song is with Shandong University and the Australian National Univeristy. X. Song is a visiting PhD student to the ANU funded by the China Scholarship Council (CSC). Email: song.sducg@gmail.com

Y. Dai is with Research School of Engineering, the Australian National University. Email: yuchao.dai@anu.edu.au

X. Qin is with School of Computer Science and Technology, Shandong Uiversity. E-mail: qxy@sdu.edu.cn

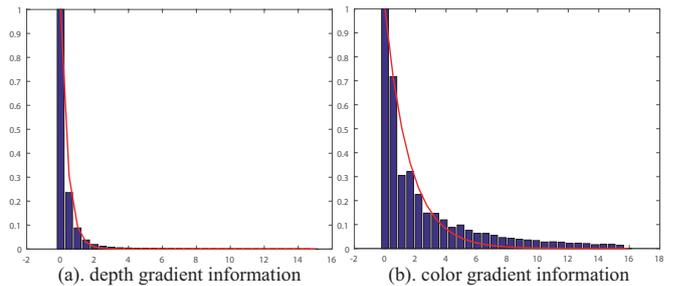

Fig. 1. Histograms of gradients of depth images and color images for the Sintel dataset [3]. (a) Histogram of gradients of depth images. (b) Histogram of gradients of color images. The distribution of both depth and color gradients can be well expressed by Laplace distribution (the red line in (a) and (b)). The $RMSE$ is 0.0016 for (a), while 0.0036 for (b). Note that depth images own sharper edges compared with the color images.

neural networks (CNN) set new records for various vision applications, especially in color image super-resolution (CSR) problems. There has been considerable progress in applying deep convolutional network for CSR and new state-of-the-art has been achieved [4] [5] [6]. It has been proven that CNN has the ability to learn the nonlinear mapping from the low resolution color images to the corresponding high resolution color images. Quite surprisingly, the success of applying CNN in color image super-resolution has not been matched to depth image super-resolution. This is mainly due to the inherent differences in the acquisition of depth image and color image. In Fig. 1, we compare the statistics of gradients of depth images and their corresponding color images (the color and depth images are already aligned). It is obvious to observe the difference between depth images and color images, where depth images generally contain less texture and sharp boundary, and are usually degraded by noise due to the imprecise consumer depth cameras (nevertheless time-of-flight or structured light) or difficulties in calculating the disparity (stereo vision).

Building upon the success of deep CNN in color image super-resolution, we propose to address the unique challenges with depth image super-resolution. First, to deal with different up-sampling ratios in depth super-resolution, we propose a progressive deep neural network structure, which gradually learns the high-frequency components and could be adapted to different depth super-resolution tasks. Second, to further refine the depth image estimation, we resort to depth field statistics and color-depth correlation to constrain the global and local structures in depth images. The corresponding problem is formulated as a conditional random field (CRF) and solved via energy minimization. Our method is elegant in dealing with or without color image in super-resolving the depth image. Extensive experiments on various bench-marking datasets demonstrate the superior performance of our method compared with current state-of-the-art methods.

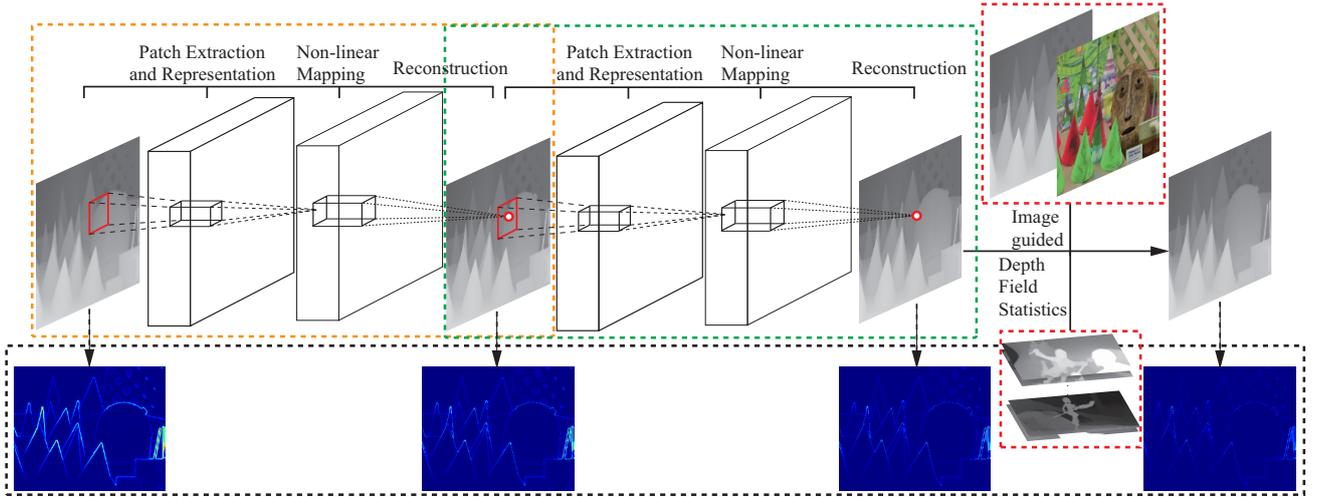

Fig. 2. Conceptual illustration of our framework, where the input is a low-resolution depth image and the output is the desired high-resolution depth image. By using convolutional neural network, our method gradually learns the high frequency components in depth images. The area labeled in black shows the residual between current depth map and the ground truth, which are shown in color mode. Blue to red means zeros to infinity. **Best Viewed on Screen.**

Our main contributions are summarized as:

1) A progressive deep CNN framework is proposed to gradually learn high-resolution depth images from low-resolution depth images. To the best of our knowledge, this is the first deep neural network based depth super-resolution method.
2) Depth field statistics and color-depth correlation are exploited to further refine the learned depth images, which demonstrates that these two priors provide complement information to the deep CNN.
3) Our method is elegant in dealing with and without high-resolution color image as guidance. For depth images without high-resolution color images, the depth images themselves can be employed to refine the depth images.

## II. Related Work

Image super-resolution, as one of the most active research topics in the field of computer vision and image processing, has been widely studied, and the research topic can be divided into two categories, namely, color image super-resolution (CSR) and depth image super-resolution (DSR).

### A. Depth image super-resolution

According to the information used, depth image super-resolution methods can be classified into the following three categories, including: depth image super-resolution from multiple depth images, single depth image super-resolution with additional depth map data-set, and depth image SR with the assistant of high resolution color image.

*a) DSR from multiple images:* Based on complimentary information among consecutive frames, [7] [8] [2] fuse multiple low-resolution depth images to obtain high resolution depth images. However, the assumption that multiple static depth images are captured with small camera movement heavily restrict the applications of these methods. Furthermore, the final results strongly depend on the camera pose estimation performance.

*b) DSR with additional depth map datasets:* Existing research has demonstrated that HR depth images can be inferred from LR depth image with prior information. Single DSR offers unique challenges compared to single CSR (*e.g.*edge preserving de-noising should also be properly tackled). Inspired by [9], Aodha et al. [10] proposed to employ a patch based MRF model in depth image super-resolution. Besides, Hornácek et al. [11] proposed to search low and high resolution patch-pairs of arbitrary size in the depth map itself. What's more, techniques based on sparse representation and dictionary learning have also been employed in DSR. Instead of using a prior information from an external database, Ferstl et al. [12] proposed to generate high-resolution depth edges by learning a dictionary of edges priors from an external database of high and low resolution examples. Then a variational energy model with Total Generalized Variation (TGV) is employed as regularization to obtain the final high resolution depth maps. However, this method suffers from blurring problems, especially areas around depth edges. What's more, Xie et al. [13] proposed to exploit a Markov random field optimization approach to generate HR depth edges from depth edges extracted from LR depth maps. Then, the HR depth edges are recognized as strong restricts to generate the final high-resolution depth maps. However, the method always loses details since it is impossible to extract all depth edges from LR depth maps. Generally, other than their own disadvantages, methods mentioned above may fail to establish patch correspondences between the external data-set and depth maps themselves, which lead to artifacts between patches or incorrect depth pattern estimation.

*c) DSR with corresponding HR color image:* Pre-aligned HR color images have also been employed to super-resolve the depth maps as the high frequency components in color images such as edges can provide useful information to assist the process of DSR. Park et al. [14] proposed a non-local means filter (NLM) method which exploits color information to maintain the detailed structure in depth maps. Yang et al. [15] proposed an adaptive color-guided auto-regressive (AR) model for high quality depth image recovery from LR depth maps. Besides, Ferstl et al. [16] utilized an anisotropic diffusion tensor to guide the depth image super-resolution. What's more, Matsuo et al. [17] described a depth image enhancement method for consumer RGB-D cameras, where color images are used as auxiliary information to compute local tangent planes in depth images. In [18], color and depth images were exploited to generate depth areas with similar color and depth information from depth images, then a sparse representation approach was employed to super-resolve depth maps in each region. However, notwithstanding the appealing results that such approaches could generate, the lack of high resolution color images fully registered with the depth images in many cases makes the color assisted approaches less general.

## B. Color image super-resolution

Effective Strategies have been proposed for color image super-resolution, which can be divided into two categories: Non-deep convolutional network (CNN) based approaches and deep convolutional network (CNN) based methods.

*a) Non-deep CNN for image super-resolution:* Firstly, Freeman et al. [9] propose an approach which formulates the super-resolution problem as a multi-class MRF model by recognizing each hidden node as the label of a high resolution patch. Unfortunately, unreliable results may be generated when no correspondence are established due to lacking of available training examples. Secondly, self-similarity has also been exploited for image super-resolution. Glasner et al. [19] propose a method searches for similar image patches across multiple down-scaled versions of the image, while the collection of training image data in advance is not needed. Huang et al. [20] employ a self-similarity strategy which expands the internal patch search space by allowing geometric variations (SRF). However, due to lacking of guarantee that patch redundancy always exists within or across image scales, acceptable results are not always achieved. Thirdly, methods based on dictionary and sparse representation of images are also widely explored. Based on the assumption that high resolution and low resolution patches should share the same reconstruction coefficients, Yang et al. [21] propose an effective method which exploits sparse linear combinations of learned coupled dictionary atoms to reconstruct high resolution and low resolution patches. Wang et al. [22] solve the super-resolution problem by relaxing the fully coupled constraint and learning a mapping through low resolution and high resolution image pairs. And similar strategies are also used in Kim et al. [23]. Moreover, Yang et al. [24] employ support vector regression (SVR) and sparse representation to obtain super-resolution results. By minimizing an error function, an SVR model is learned to reconstruct the high resolution images. Neither collection of low and high resolution training data, nor the prior information of self-similarity is needed.

*b) Deep CNN for image super-resolution:* Recently, deep CNN has extended its success in high level computer vision to low level computer vision tasks such as color image super-resolution. Dong et al. [4] proposed an end-to-end deep CNN framework to learn the nonlinear mapping between low and high resolution images. Based on [4], Kim et al. [5] proposed to use a deeper network to represent the non-linear mapping and improved performance has been achieved. Meanwhile, in [6], a deeply-recursive convolutional network for color images super-resolution is proposed. Based on a deep recursive layer, a more accurate representation of the mapping between low and high resolution color images can be achieved, thus, better results can be generated without introducing new parameters.

## III. OUR APPROACH

In this paper, build upon the success of deep CNN for color image super-resolution, we propose a deep neural network based depth image super-resolution method to effectively learn the non-linear mapping from low-resolution depth images to high-resolution depth images. Furthermore, we enforce the depth field statistics and local color-depth correlation to further refine the results.

Depth image super-resolution aims at inferring a high-resolution depth image $\mathbf{D}^H$ from a low-resolution depth image $\mathbf{D}^L$, where the up-sampling factors vary from $\times 2$, $\times 4$ to $\times 8$ (infer $8 \times 8 = 64$ depth values from a single depth value). Due to the information loss in subsampling and the nonlinear mapping between low resolution depth images and high-resolution depth images, this is a very challenging task. To effectively learn the nonlinear mapping $\mathbf{F}$ from low-resolution depth image $\mathbf{D}^L$ to high-resolution depth image $\mathbf{D}^H$, we propose to use deep neural network. To deal with different up-sampling factors, our proposed deep neural network could gradually learn the high frequency component in depth images and in this way progressively improve the performance. By using this progressive structure, we are able to modify the network structure adaptively for different super-resolution tasks. To further constrain the high-resolution depth images, we resort to the depth field statistics and local color-depth correlation priors, which results in an energy minimization formulation.

Fig. 2 provides a conceptual illustration of our proposed framework, which consists of the progressive deep neural network, depth field statistics prior and color-depth correlation prior. Given a low-resolution depth image as input (interpolated to a high-resolution depth image by bicubic firstly), our method outputs a corresponding high-resolution depth image through deep neural network learning and energy minimization. In Fig. 3, we compare the output at different stages of our system. It could be observed that the data goes from the left side to the right side, quality of the high-resolution depth image gradually improves (as indicated by RMSE), which demonstrates that our method gradually learns the fine details in the depth image.

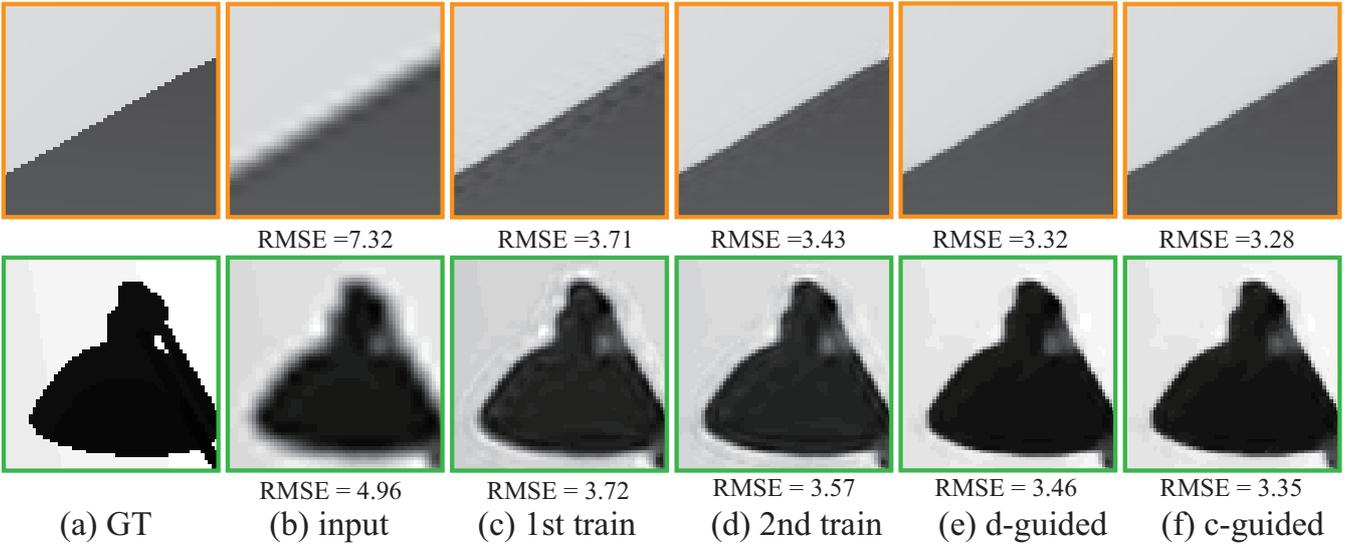

Fig. 3. Typical outputs at different stages of of our method. (a) Ground truth high-resolution depth image, (b) Input low-resolution depth map (interpolated to the same size), (c) Output after the first nonlinear mapping, (d) Output after the second nonlinear mapping, (e) and (f) show the depth super-resolution results with depth field statistics and color-depth correlation correspondingly. **Best Viewed on Screen.**

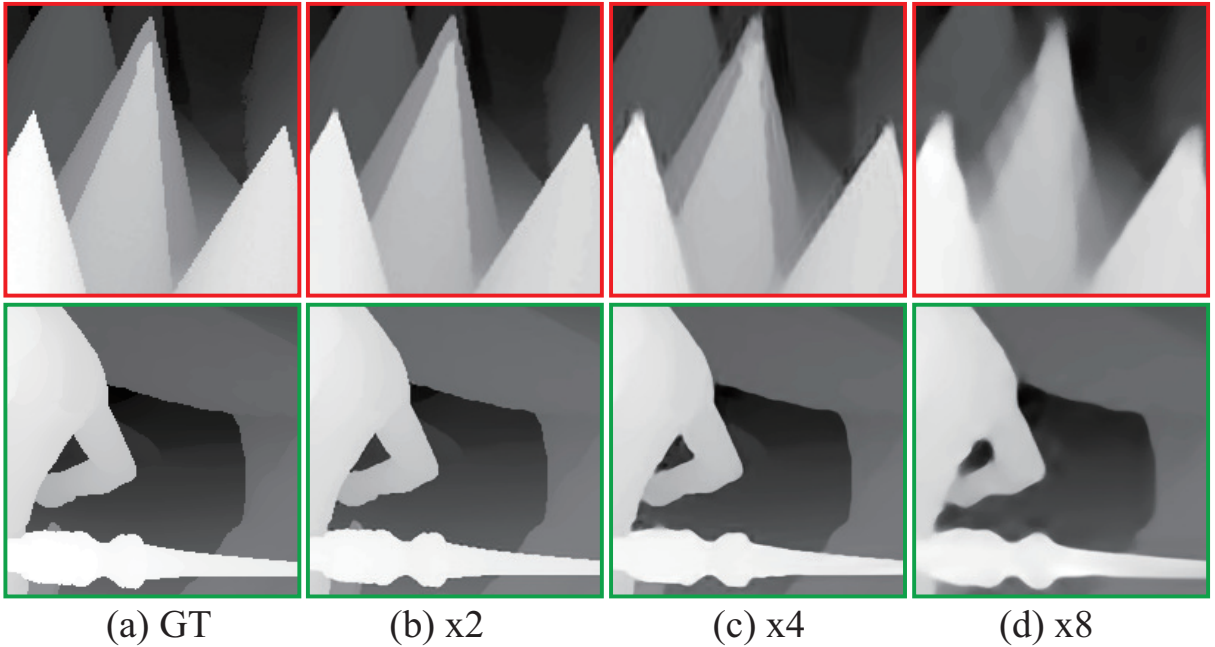

Fig. 4. (a) Ground truth, (b) results of ×2, (c) results of ×4, (d) results of ×8. The first row show the part of images extracted from Middlebury-set, and the second row shows the part of images extracted from Sintel data-set.

*A. Learning depth super-resolution by deep CNN*

Deep Convolutional Neural Networks (DCNNs) have recently shown the state of the art performance in high level vision tasks, such as image classification and object detection. This work brings together methods from DCNNs and probabilistic graphical models to address the task of depth image super-resolution. High-resolution depth image relates the corresponding low-resolution one by a non-linear mapping due to the sub-sampling. The complexity of the non-linear mapping depends on the up-sampling factor as illustrated in Fig. 4. To effectively capture the high-frequency details in the depth images, we propose a progressive CNN structure as illustrated in Fig. 2. The progressive CNN structure consists of layers of depth super-resolution units. Each unit maps the low-resolution input $\mathbf{D}_i^L$ to the high-resolution output $\mathbf{D}_{i+1}^L$, which is the input of the next unit. In this way, our network is capable of gradually learning the high-frequency components in high-resolution depth images. As shown in the black labeled area in figure. 2, we can see that the residual reduces, which means that high-frequency components can be learned in each step.

*a) Low-resolution to high-resolution mapping unit:* As shown in Fig. 2 (areas labeled in green and orange), for each low-resolution to high-resolution learning unit in our progressive CNN structure, we adopt the same network structure from Dong *et.al.* [4]. It is worth noting that other network structures such as [5], [6] could also be used. The unit consists of three operators, namely, patch extraction and representation, non-linear mapping and reconstruction, which together learn the mapping $\mathbf{F}$ to map the input low resolution images $\mathbf{D}^L$ to a high-resolution one $\mathbf{F}(\mathbf{D}^L)$. Note that, we train the network from scratch rather than fine tune the network from [4] due to the inherent differences between depth images and color images. In principle, we could train the progressive network together from the low-resolution input to the desired high-resolution output. For the ease of training, we instead train each unit consecutively. In this way, the training procedure has been greatly simplified. A final fine tune of the whole progressive network could be implemented.

*b) Implementation Details:* 128 depth maps extracted from the Middlebury stereo dataset, Sintel dataset and synthesis depth maps are exploited to construct the training data to train our network, and 14 depth maps are used as test images. 128 depth image dataset can be decomposed into 255000 sub-images, which are extracted from original depth images with a stride of 14. And the size of each sub-image is $33 \times 33$. For Patch Extraction and Representation, 64 convolutions are employed, and each convolution has a kernel size of $9 \times 9$. For Non-linear Mapping, 32 convolutions are used with the kernel size $1 \times 1$ in each convolution. For Reconstruction, the kernel size is $5 \times 5$. Note that we first upscale the low-resolution depth image to the desired size using bicubic interpolation, and the upscaled depth image is our initial input.

### B. Color modulated smoothness for depth super-resolution

In the above section, we have demonstrated how high-resolution depth images can be learned from low-resolution depth images by using *progressive convolutional neural network*, which already achieve comparable performance with the current state-of-the-art methods [12] [13]. However, as the depth super-resolution learning is achieved in an end-to-end framework and learned across a dataset, the learned depth image could be biased by different RGB-D statistics (as illustrated by Fig. 4). As this depth map is learned from a dataset, we cannot expect the network to learn the high-frequency information or edge. Therefore we resort to two different depth super-resolution cues, which are in principle complement to the above end-to-end learning, namely, depth field statistics and color image guidance. What is interesting is that we show that when the color image guidance is not available, the depth image itself could be used as guidance.

Color modulated smoothness term aims at representing the local structure of the depth map. And depth maps for generic 3D scenes contain mainly smooth regions separated by curves with sharp boundaries. The key insight behind the color modulated smoothness term is that the depth map and the color image are locally correlated, thus the local structure of the depth map can be inferred with the guidance of the corresponding color image. The term has already employed in image colorization [28], depth in-painting [29] and depth image super resolution [30] [10].

Denote $D_u$ as the depth value at location $u$ in a depth image, the depth image inferred by the model can be expressed as:

$$D_u = \sum_{v \in \theta_u} \alpha_{(u,v)} D_v, \qquad (1)$$

where $\theta_u$ is the neighborhood of pixel $u$ and $\alpha_{(u,v)}$ denotes the color modulated smoothness model coefficient for pixel $v$ in the set of $\theta_u$. The discrepancy between the model and the depth map (*i.e.*, the color modulated smoothness potential) can be expressed as:

$$\psi_\theta(\mathbf{D}_\theta) = \left( D_u - \sum_{r \in \theta_u} \alpha_{(u,v)} D_v \right)^2. \qquad (2)$$

The color modulated smoothness term defined on the whole image can be expressed as:

$$\sum_u \psi_\theta(\mathbf{D}_\theta) = \|\mathbf{A}\mathrm{vec}(\mathbf{D}) - \mathbf{b}\|^2, \qquad (3)$$

where $\mathbf{A}$ and $\mathbf{b}$ are obtained from $\alpha_{u,v}$.

We need to design a local color modulated smoothness predictor $\alpha_{(u,v)}$ with the available high-resolution color image. To avoid incorrect depth prediction due to depth-color inconsistency, depth information is also included in $\alpha$, hence,

$$\alpha_{(u,v)} = \frac{1}{N_u} \alpha_{(u,v)}^{\overline{D}} \alpha_{(u,v)}^{I}, \qquad (4)$$

where $N_u$ is the normalization factor, $\overline{D}$ is the observed depth map (generated by our progressive deep CNN), $I$ is the corresponding high-resolution color image, $\alpha_{(u,v)}^{\overline{D}} \propto \exp(-(\overline{D}_u - \overline{D}_v)^2 / 2\sigma_{\overline{D}_u}^2)$, and $\alpha_{(u,v)}^{I} \propto \exp(-(g_u - g_v)^2 / 2\sigma_{I_u}^2)$, where $g$ represents the intensity value of corresponding color pixels, $\sigma_{\overline{D}_u}$ and $\sigma_{I_u}$ are the variance of the depth and color intensities in the local path around $u$.

The window size for local neighborhood $\theta$ is set as $7 \times 7$ in our experiment. Note that for a depth map without corresponding color image, the depth map itself can be used to build $\alpha$.

### C. Depth super-resolution cue from depth filed statistics

The distribution of a natural depth image $\mathbf{D}$ can often be modeled as a generalized Laplace distribution (a.k.a., generalized Gaussian distribution). As illustrated in Fig. 1, the distribution of gradient of depth images can be well approximated with Laplacian distribution. Therefore, we propose to minimize the total variation of the depth image, i.e. $\|\mathbf{D}\|_{\mathrm{TV}} \to \min$, where the total variation could be expressed in matrix form as:

$$\|\mathbf{D}\|_{\mathrm{TV}} = \|\mathbf{P}\mathrm{vec}(\mathbf{D})\|_1, \qquad (5)$$

where $\mathbf{P}$ extracts the gradients along $X$ and $Y$ directions.

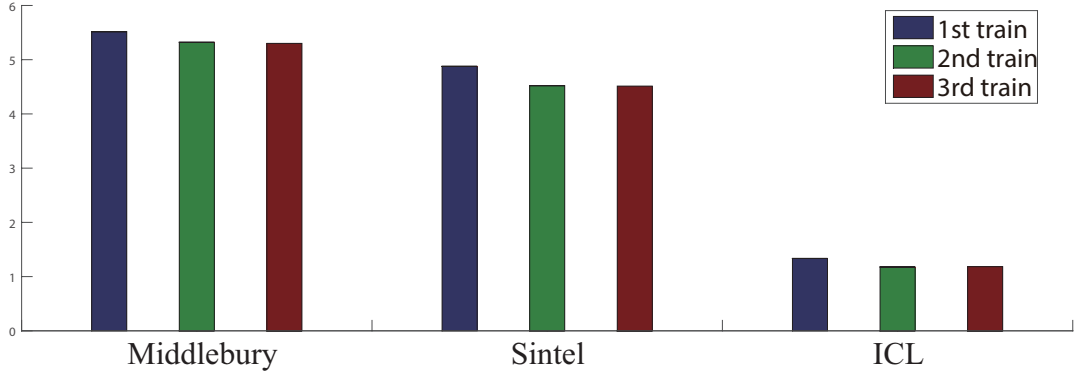

Fig. 5. The figure shows the comparison of our multi-train $RMSE$ results for upsampling factor of $\times 8$, from which, we can see that the $RMSE$ of results generated by 2nd train is much smaller than the 1st train, while the 3rd train is similar with the 2nd train.

*D. An Energy Minimization Formulation*

To further refine the depth super-resolution results, we integrate the depth super-resolution cue from deep progressive CNN, depth field statistics and color/depth guidance and reach the following energy minimization formulation:

$$\min_{\mathbf{D}} \frac{1}{2}\|\mathbf{D}-\overline{\mathbf{D}}\|_F^2 + \lambda_1\|\mathbf{A}\text{vec}(\mathbf{D})-\mathbf{b}\|^2 + \lambda_2\|\mathbf{P}\text{vec}(\mathbf{D})\|_1, \quad (6)$$

where $\overline{\mathbf{D}}$ is the depth super-resolution result from the deep neural network, $\mathbf{A}$ and $\mathbf{b}$ express the color/depth modulated smoothness constraint, and $\mathbf{D}$ is the final depth map we want to generate.

This is a convex optimization problem where a global optimal solution exists. Here we propose to use iterative reweighted least squares (IRLS) to efficiently solve the problem. Given the depth map estimation result in the $it$-th iteration, the optimization for the $it+1$-th iteration is expressed as:

$$\begin{aligned}
&\min_{\mathbf{D}} \frac{1}{2}\|\mathbf{D}-\overline{\mathbf{D}}\|_F^2 + \lambda_1\|\mathbf{A}\text{vec}(\mathbf{D})-\mathbf{b}\|^2 \\
&\qquad + \lambda_2 \sum_i \|\mathbf{P}_i\text{vec}(\mathbf{D})\| \\
&= \min_{\mathbf{D}} \frac{1}{2}\|\mathbf{D}-\overline{\mathbf{D}}\|_F^2 + \lambda_1\|\mathbf{A}\text{vec}(\mathbf{D})-\mathbf{b}\|^2 + \\
&\qquad \lambda_2 \sum_i \frac{\|\mathbf{P}_i\text{vec}(\mathbf{D})\|^2}{\|\mathbf{P}_i\text{vec}(\mathbf{D}^{(it)})\|},
\end{aligned} \quad (7)$$

which could be equivalently expressed as:

$$\min_{\mathbf{D}} \frac{1}{2}\|\mathbf{D}-\overline{\mathbf{D}}\|_F^2 + \lambda_1\|\mathbf{A}\text{vec}(\mathbf{D})-\mathbf{b}\|^2 + \\ \lambda_2 \sum_i \|\frac{\mathbf{P}_i}{\sqrt{\|\mathbf{P}_i\text{vec}(\mathbf{D}^{(it)})\|}}\text{vec}(\mathbf{D})\|^2. \quad (8)$$

Denote $\mathbf{E}_i^{(it)} = \frac{\mathbf{P}_i}{\sqrt{\|\mathbf{P}_i\text{vec}(\mathbf{D}^{(it)})\|}}$, i.e., the row-wise reweighted version of $\mathbf{P}_i$ and $\mathbf{E}^{(it)} = \left[\mathbf{E}_i^{(it)}\right]$, we have:

$$\mathbf{D}^{(it+1)} = \arg\min_{\mathbf{D}} \frac{1}{2}\|\mathbf{D}-\overline{\mathbf{D}}\|_F^2 + \lambda_1\|\mathbf{A}\text{vec}(\mathbf{D})-\mathbf{b}\|^2 + \\ \lambda_2\|\mathbf{E}^{(it)}\text{vec}(\mathbf{D})\|_F^2. \quad (9)$$

The above least squares problem owns a closed-form solution.

We apply iterative reweighted least squares (IRLS) [32] [33] to efficiently solve the above energy minimization problem. $\lambda_1$ and $\lambda_2$ are set as $0.7$ empirically in our experiment.

## IV. EXPERIMENTS

In this section we evaluate the performance of the different state-of-the-art ($SOTA$) methods on publicly available datasets, including Middlebury stereo dataset [34] [35] [36] [37], the Laser Scan dataset provided by Aodha et al. [10], Sintel dataset [3] and ICL dataset [38]. The Middlebury data-set is the most widely used stereo data-set and is also the data-set most super-resolution methods use. This data-set contains high resolution depth with lots of details and the images contain lots of textures, which are relatively challenging for segmentation. Moreover, the Sintel data-set is a synthesized data-set and contains lots of depth details and high quality images. This data-set uses physical simulation to synthesize complex scenes.

*a) Error metrics:* In this paper, we use three kind of error metrics to evaluate the results obtained by our method and other state-of-the-art methods, including:

1) Root Mean Squared Error ($RMSE$) of recovered depth image with respect to the ground truth depth image, *i.e.*

$$RMSE = \sqrt{\sum_{i=1}^{N}(X_{obs,i}-X_{gt,i})^2/N}, \quad (10)$$

where $X_{obs}$ and $X_{gt}$ are the observe data and ground truth respectively. $RMSE$ represents the sample standard deviation of the differences between predicted values and Ground truth data.

2) Mean Absolute Error($MAE$) of recovered depth with respect to the ground truth depth image,

$$MAE = \sum_{i=1}^{N} abs(X_{obs,i}-X_{gt,i})/N. \quad (11)$$

3) Structure Similarity of Index ($SSIM$) of the recovered depth with respect to the ground truth depth image.

Generally, we obtain high resolution depth maps, subsample them, reconstruct using our methods, then compare the reconstruction to the original depth maps. For the Sintel data-set, the high resolution depth maps are ground truth; for others, they are the best available depth maps. We aim to produce reconstructions that are as close as possible to the high resolution depth images.

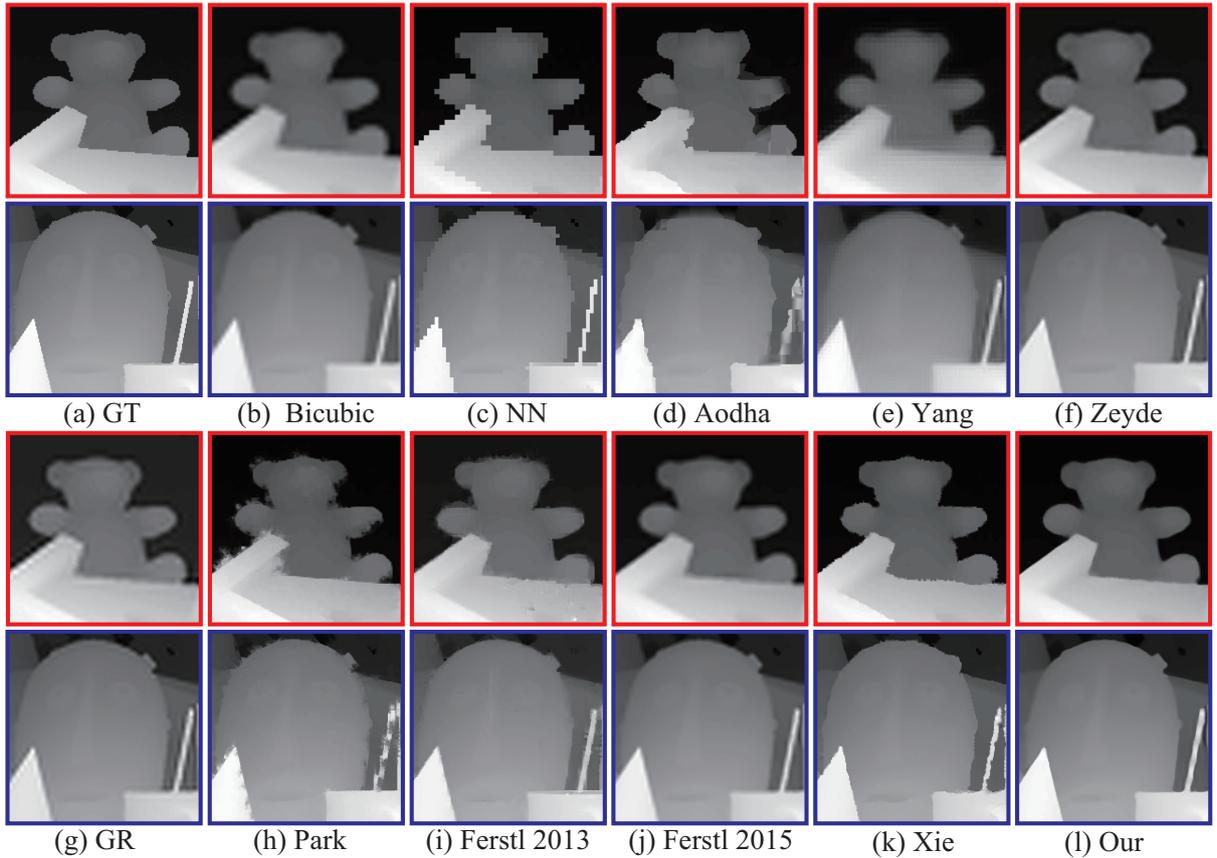

Fig. 6. Middlebury results (factor ×4). (a) Ground Truth. (b) Bicubic method. (c) Nearest neighbor method. (d) Aodha et al. [10]. (e) Yang et al. [25]. (f) Zeyde et al. [26]. (g) GR [27]. (h) Park et al. [14]. (i) Ferstl et al. [16]. (j) Ferstl et al. [12]. (k) Xie et al. [13]. (l) Our method. **Best viewed on Screen.**

*b) Parameters Setting:* We set the same parameters in all of our experiments for different scales and different images via trial-and-error.

*c) Baseline Methods:* We compare our results with the following three categories of the methods. 1) State-of-the-art single depth image super resolution methods, including Aodha et al. [10], Hornacek et al.[11], Ferstl et al. [12] and Xie et al. [13]; 2) state-of-the-art color assisted depth image super resolution approaches, including Park et al. [14], Yang et al. [31], Ferstl et al. [16]; 3) General single color image super resolution approaches, including Zeyde et al.[26], Yang et al.[25] and Timofte et al. [27], which contains two kinds of methods: Global Regression ($GR$) and Anchored Neighborhood Regression ($ANR$), the neighborhood embedding methods proposed by Bevilacqua et al. [39], which contains $NE + LS$, $NE + NNLS$ and $NE + LLE$, and Huang et al. [20]; 4) Standard interpolate approaches, including Bicubic and Nearest Neighbour ($NN$). We either use the source code provided by the authors or implement those methods by ourselves. We also select the parameters of baseline methods by experiments. More specifically, we use a few sets of parameters settings for each method (at least one of them is from the default parameters provided by the authors and the others are chosen by ourselves) and we choose the one that generates the lowest percentage of errors for all the testing images.

A. Analysis

In this section, we present experimental analysis to the contribution of each component of our method, namely, deep regression, depth field statistics and color (or depth) guidance. Figure 3 (parts of images from $Sintel$ (first row) and $ICL$ (second row) data-set) shows the visual comparison in different stages of our method. We can see clearly that 1st trained results are much better than the input data (obtained by bicubic), 1st trained results not only have much smaller $RMSE$, but also have better visualization, which means that high frequency depth information can be learned by the deep convolutional network (SRCNN) in this step effectively. However, the 1st trained results are not good enough and suffer from ring effect severely. And, as shown in figure 3 (d), our 2nd trained network generates better results which generates sharper boundaries and reduces the influences of ring effect. It is obviously to find that high frequency depth information can also be learned in this step. Based on figure. 3 (f), we can see that depth field statistics and the guided color images can

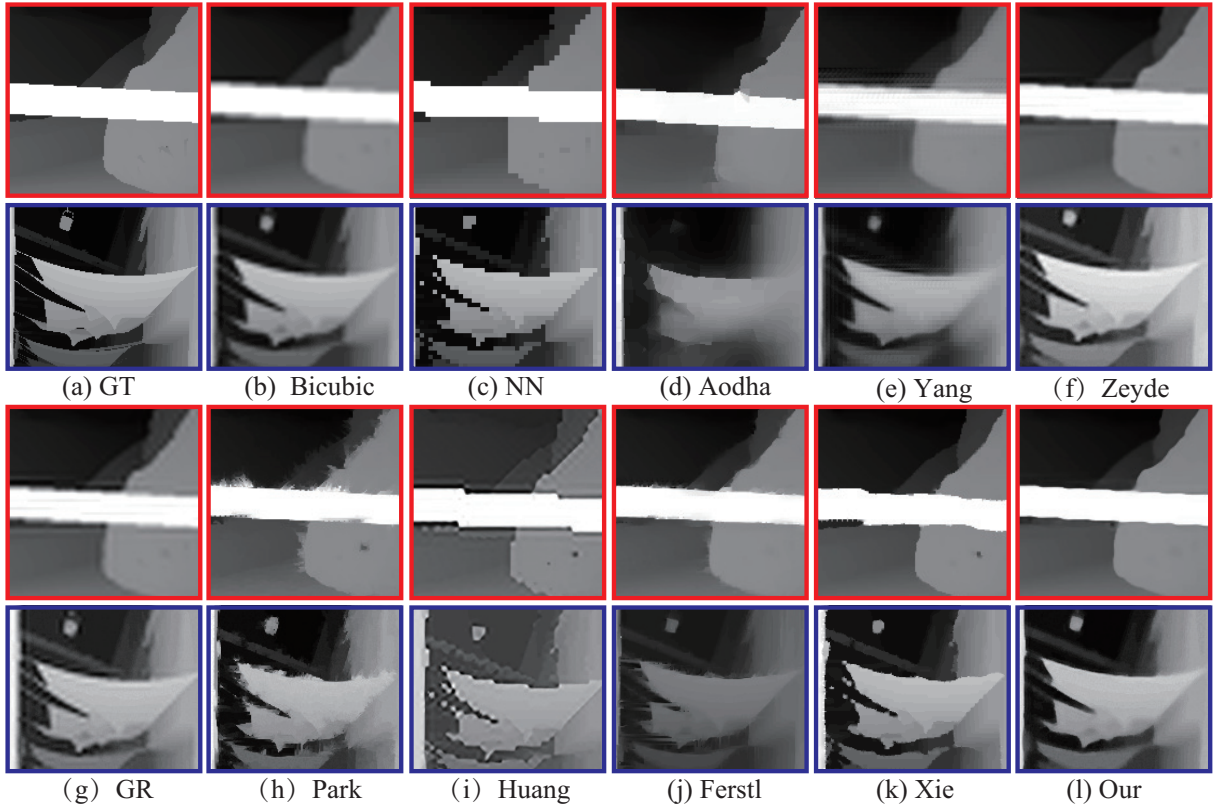

Fig. 7. Sintel results (factor ×4). (a) Ground Truth. (b) Bicubic. (c) Nearest neighbor. (d) Aodha et al. [10]. (e) Yang et al. [25]. (f) Zeyde et al. [26]. (g) GR [27]. (h) Park et al. [14]. (i) Huang et al. [20]. (j) Ferstl et al. [16]. (k) Xie et al. [13]. (l) Our method. **Best viewed on screen**

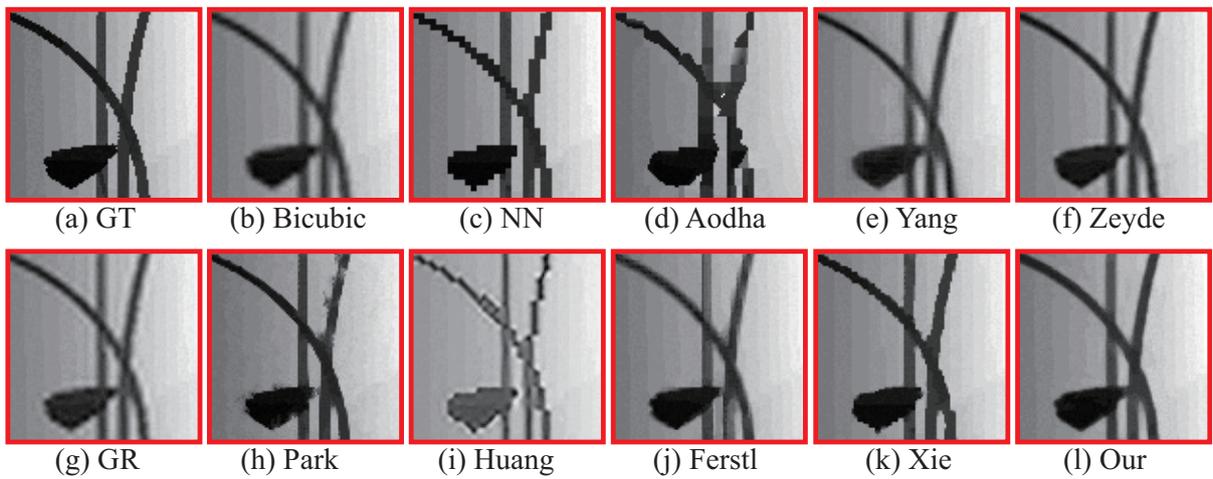

Fig. 8. ICL result (factor ×3). (a) Ground Truth. (b) Bicubic. (c) Nearest neighbor. (d) Aodha et al. [10]. (e) Yang et al. [25]. (f) Zeyde et al. [26]. (g) GR [27]. (h) Park et al. [14]. (i) Huang et al. [20]. (j) Ferstl et al. [16]. (k) Xie et al. [13]. (l) Our method. **Best viewed on screen.**

## TABLE I

| Method | ×2 | | | | ×4 | | | | ×4 | | |
|---|---|---|---|---|---|---|---|---|---|---|---|
| | Cones | Teddy | Tsukuba | Venus | Cones | Teddy | Tsukuba | Venus | Scan21 | Scan30 | Scan42 |
| NN | 4.4622 | 3.2363 | 9.2305 | 2.1298 | 6.0054 | 4.5466 | 12.9083 | 2.9333 | 2.6474 | 2.5196 | 5.6044 |
| Bicubic | 2.5245 | 1.9495 | 5.7828 | 1.3119 | 3.8635 | 2.8930 | 8.7103 | 1.9403 | 2.0324 | 1.9764 | 4.5813 |
| Park et al. [14] | 2.8497 | 2.185 | 6.8869 | 1.2584 | 6.5447 | 4.3366 | 12.1231 | 2.2595 | N/A | N/A | N/A |
| Yang et al. [31] | 2.4214 | 1.8941 | 5.6312 | 1.2368 | 5.1390 | 4.0660 | 13.1748 | 2.7559 | N/A | N/A | N/A |
| Ferstl et al. [16] | 3.1651 | 2.4208 | 6.9988 | 1.4194 | 3.9968 | 2.8080 | 10.0352 | 1.6643 | N/A | N/A | N/A |
| Yang et al. [25] | 2.8384 | 2.0079 | 6.1157 | 1.3777 | 3.9546 | 3.0908 | 8.2713 | 1.9850 | 2.0885 | 2.0349 | 4.7474 |
| Zeyde et al. [26] | 1.9539 | 1.5013 | <u>4.5276</u> | <u>0.9305</u> | <u>3.2232</u> | 2.3527 | 7.3003 | 1.4751 | 1.6869 | 1.6207 | 3.6414 |
| GR [27] | 2.3742 | 1.8010 | 5.4059 | 1.2153 | 3.5728 | 2.7044 | 8.0645 | 1.8175 | 1.8906 | 1.8462 | 3.9806 |
| ANR [27] | 2.1237 | <u>1.6054</u> | 4.8169 | 1.0566 | 3.3156 | 2.4861 | 7.4895 | 1.6449 | 1.7334 | 1.6823 | 3.8140 |
| NE+LS | <u>2.0437</u> | 1.5256 | 4.6372 | 0.9697 | 3.2868 | <u>2.4210</u> | <u>7.3404</u> | 1.5225 | 8.6852 | 8.8460 | 8.1874 |
| NE+NNLS | 2.1158 | 1.5771 | 4.7287 | 1.0046 | 3.4362 | 2.4887 | 7.5344 | <u>1.6291</u> | 1.7313 | 1.6849 | 3.5733 |
| NE+LLE | 2.1437 | 1.6173 | 4.8719 | 1.0827 | 3.3414 | 2.4905 | 7.5528 | 1.6449 | 1.7058 | 1.6547 | 3.7975 |
| Aodha et al. [10] | 4.3185 | 3.2828 | 9.1089 | 2.2098 | 12.6938 | 4.1113 | 12.6938 | 2.6497 | 2.5983 | 2.6267 | 6.1871 |
| Hornácek et al. [11] | 3.7512 | 3.1395 | 8.8070 | 2.0383 | 5.4898 | 5.0212 | 11.1101 | 3.5833 | 2.8585 | 2.7243 | 4.5074 |
| Huang et al. [20] | 4.6273 | 3.4293 | 10.0766 | 2.1653 | 6.2723 | 4.8346 | 13.7645 | 3.0606 | 2.7097 | 2.6245 | 5.9896 |
| Ferstl et al. [12] | 2.2139 | 1.7205 | 5.3252 | 1.1230 | 3.5680 | 2.6474 | 7.5356 | 1.7771 | 1.4349 | 1.4298 | 3.1410 |
| Xie et al. [13] | 2.7338 | 2.4911 | 6.3534 | 1.6390 | 4.4087 | 3.2768 | 9.7765 | 2.3714 | <u>1.3993</u> | <u>1.4101</u> | <u>2.6910</u> |
| Our | **1.4356** | **1.1974** | **2.9841** | **0.5592** | **2.9789** | **1.8006** | **6.1422** | **0.8796** | **1.1135** | **1.0711** | **1.6658** |

Quantitative evaluation. The $RMSE$ is calculated for different SOTA methods for the Middlebury and Laserscanner data-sets for factors of ×2 and ×4. The best result of all single image methods for each data-set and up-scaling factor is highlighted and the second best is underlined.

## TABLE II

| Method | ×2 | | | | ×4 | | | | ×4 | | |
|---|---|---|---|---|---|---|---|---|---|---|---|
| | Cones | Teddy | Tsukuba | Venus | Cones | Teddy | Tsukuba | Venus | Scan21 | Scan30 | Scan42 |
| NN | 0.9645 | 0.9696 | 0.9423 | 0.9888 | 0.9360 | 0.9450 | 0.9003 | 0.9800 | 0.9814 | 0.9828 | 0.9679 |
| Bicubic | 0.9720 | 0.9771 | 0.9536 | 0.9909 | 0.9538 | 0.9619 | 0.9205 | 0.9845 | 0.9875 | 0.9879 | 0.9743 |
| Park et al. [14] | 0.9699 | 0.9767 | 0.9320 | 0.9910 | 0.9420 | 0.9553 | 0.8981 | 0.9862 | N/A | N/A | N/A |
| Yang et al. [31] | 0.9831 | 0.9851 | 0.9720 | 0.9944 | 0.9624 | 0.9695 | 0.9314 | <u>0.9879</u> | N/A | N/A | N/A |
| Ferstl et al. [16] | 0.9755 | 0.9795 | 0.9576 | 0.9938 | 0.9625 | 0.9707 | 0.9245 | <u>0.9901</u> | N/A | N/A | N/A |
| Yang et al. [25] | 0.9473 | 0.9564 | 0.9072 | 0.9805 | 0.9482 | 0.9566 | 0.9014 | 0.9816 | 0.9860 | 0.9866 | 0.9690 |
| Zeyde et al. [26] | 0.9655 | 0.9717 | 0.9438 | 0.9886 | 0.9604 | 0.9628 | 0.9147 | 0.9883 | 0.9908 | 0.9912 | 0.9830 |
| GR [27] | 0.9587 | 0.9656 | 0.9314 | 0.9862 | 0.9500 | 0.9592 | 0.9012 | 0.9817 | 0.9880 | 0.9885 | 0.9763 |
| ANR [27] | 0.9630 | 0.9693 | 0.9400 | 0.9879 | 0.9391 | 0.9452 | 0.8731 | 0.9806 | 0.9895 | 0.9898 | 0.9796 |
| NE+LS | 0.9623 | 0.9692 | 0.9391 | 0.9887 | 0.9514 | 0.9574 | 0.9042 | 0.9852 | 0.9355 | 0.9182 | 0.9302 |
| NE+NNLS | 0.9640 | 0.9707 | 0.9426 | 0.9883 | 0.9424 | 0.9499 | 0.8872 | 0.9820 | 0.9896 | 0.9900 | 0.9805 |
| NE+LLE | 0.9588 | 0.9658 | 0.9405 | 0.9837 | 0.9270 | 0.9331 | 0.8794 | 0.9641 | 0.9896 | 0.9896 | 0.9783 |
| Aodha et al. [10] | 0.9606 | 0.9690 | 0.9364 | 0.9874 | 0.9392 | 0.9520 | 0.9080 | 0.9822 | 0.9838 | 0.9838 | 0.9668 |
| Hornácek et al. [11] | 0.9696 | 0.9719 | 0.9461 | 0.9895 | 0.9501 | 0.9503 | 0.9137 | 0.9789 | 0.9814 | 0.9825 | 0.9754 |
| Huang et al. [20] | 0.9582 | 0.9673 | 0.9301 | 0.9875 | 0.9360 | 0.9425 | 0.8821 | 0.9784 | 0.9808 | 0.9819 | 0.9602 |
| Ferstl et al. [12] | <u>0.9866</u> | <u>0.9884</u> | <u>0.9766</u> | <u>0.9963</u> | <u>0.9645</u> | <u>0.9716</u> | <u>0.9413</u> | 0.9893 | <u>0.9918</u> | <u>0.9916</u> | 0.9819 |
| Xie et al. [13] | 0.9633 | 0.9625 | 0.9464 | 0.9852 | 0.9319 | 0.9331 | 0.8822 | 0.9730 | 0.9869 | 0.9878 | <u>0.9899</u> |
| Our | **0.9989** | **0.9918** | **0.9905** | **0.9989** | **0.9783** | **0.9831** | **0.9666** | **0.9973** | **0.9948** | **0.9947** | **0.9939** |

Quantitative evaluation. The $SSIM$ is calculated for different SOTA methods for the Middlebury and Laserscanner data-sets for factors of ×2 and ×4. The best result of all single image methods for each data-set and up-scaling factor is highlighted and the second best is underlined.

## TABLE III

| Method | ×2 | | | | ×4 | | | | ×4 | | |
|---|---|---|---|---|---|---|---|---|---|---|---|
| | Cones | Teddy | Tsukuba | Venus | Cones | Teddy | Tsukuba | Venus | Scan21 | Scan30 | Scan42 |
| NN | 0.6041 | 0.5133 | 0.9977 | 0.1992 | 1.1320 | 0.9117 | 1.9156 | 0.3767 | 0.4095 | 0.3963 | 0.6986 |
| Bicubic | 0.8330 | 0.6541 | 1.5719 | 0.2713 | 1.3576 | 1.0315 | 2.7140 | 0.4646 | 0.4240 | 0.4036 | 0.8332 |
| Park et al. [14] | 0.6826 | 0.5661 | 1.3345 | 0.2232 | 1.8421 | 1.3585 | 3.0010 | 0.5373 | N/A | N/A | N/A |
| Yang et al. [31] | 0.6624 | 0.5310 | 1.2812 | 0.2432 | 1.2275 | 0.9172 | 2.6021 | 0.4298 | N/A | N/A | N/A |
| Ferstl et al. [16] | 0.7209 | 0.5920 | 1.3745 | 0.2043 | 1.1456 | 0.8344 | 2.6304 | 0.3344 | N/A | N/A | N/A |
| Yang et al. [25] | 1.497 | 1.1426 | 3.0513 | 0.5061 | 1.6424 | 1.2716 | 3.1164 | 0.5901 | 0.5242 | 0.4883 | 1.0317 |
| Zeyde et al. [26] | 0.8996 | 0.7108 | 1.6921 | 0.3117 | 1.4298 | 1.1858 | 3.0439 | 0.4718 | 0.3253 | 0.3131 | 0.5744 |
| GR [27] | 1.1522 | 0.8966 | 2.2505 | 0.3933 | 1.9623 | 1.4814 | 3.8631 | 0.7245 | 0.4160 | 0.3950 | 0.8137 |
| ANR [27] | 0.9920 | 0.7925 | 1.8909 | 0.3455 | 1.9852 | 1.6239 | 4.1730 | 0.7244 | 0.3894 | 0.3730 | 0.7251 |
| NE+LS | 0.9681 | 0.7616 | 1.8311 | 0.2967 | 1.6977 | 1.3322 | 3.3549 | 0.6156 | 1.5811 | 1.8025 | 1.7138 |
| NE+NNLS | 0.9735 | 0.7574 | 1.7840 | 0.3217 | 1.8210 | 1.5839 | 3.7954 | 0.6436 | 0.3753 | 0.3589 | 0.6881 |
| NE+LLE | 1.1624 | 0.9396 | 1.8720 | 0.5256 | 2.7040 | 2.3083 | 4.2694 | 1.6085 | 0.4142 | 0.4225 | 0.8515 |
| Aodha et al. [10] | 0.8242 | 0.6549 | 1.4446 | 0.2676 | 1.5674 | 1.0883 | 2.8786 | 0.4274 | 0.4436 | 0.4388 | 0.8931 |
| Hornácek et al. [11] | 0.6310 | 0.5620 | 1.1330 | 0.2052 | 1.0432 | 0.9959 | 1.9524 | 0.7475 | 0.4290 | 0.4253 | 0.5566 |
| Huang et al. [20] | 1.0548 | 0.8333 | 2.0629 | 0.3533 | 1.8642 | 1.4754 | 3.8597 | 0.6218 | 0.5973 | 0.5644 | 1.1792 |
| Ferstl et al. [12] | 0.5259 | <u>0.3900</u> | 1.0601 | <u>0.1223</u> | 1.1034 | 0.7902 | 2.1873 | <u>0.3000</u> | 0.3099 | 0.3054 | 0.6487 |
| Xie et al. [13] | <u>0.4612</u> | 0.4238 | <u>0.7074</u> | 0.2067 | <u>0.9620</u> | <u>0.7717</u> | <u>1.5937</u> | 0.3659 | **0.2312** | **0.2372** | <u>0.3348</u> |
| Our | **0.3849** | **0.3167** | **0.5605** | **0.0641** | **0.7800** | **0.5315** | **1.4346** | **0.1270** | <u>0.2581</u> | <u>0.2401</u> | **0.3225** |

Quantitative evaluation. The $MAE$ is calculated for different SOTA methods on the Middlebury and Laserscanner data-sets for factors of ×2 and ×4. The best result of all single image methods for each dataset and up-scaling factor is highlighted and the second best is underlined.

further improve the quality of depth maps, sharper boundaries are generated and the influences of ring effect are almost eliminated. As shown in figure. 3 (e), it is also interesting to see that depth maps themselves can be recognized as guidance to further improve the qualities of depth maps. What's more, color image guided results are better than depth image guided

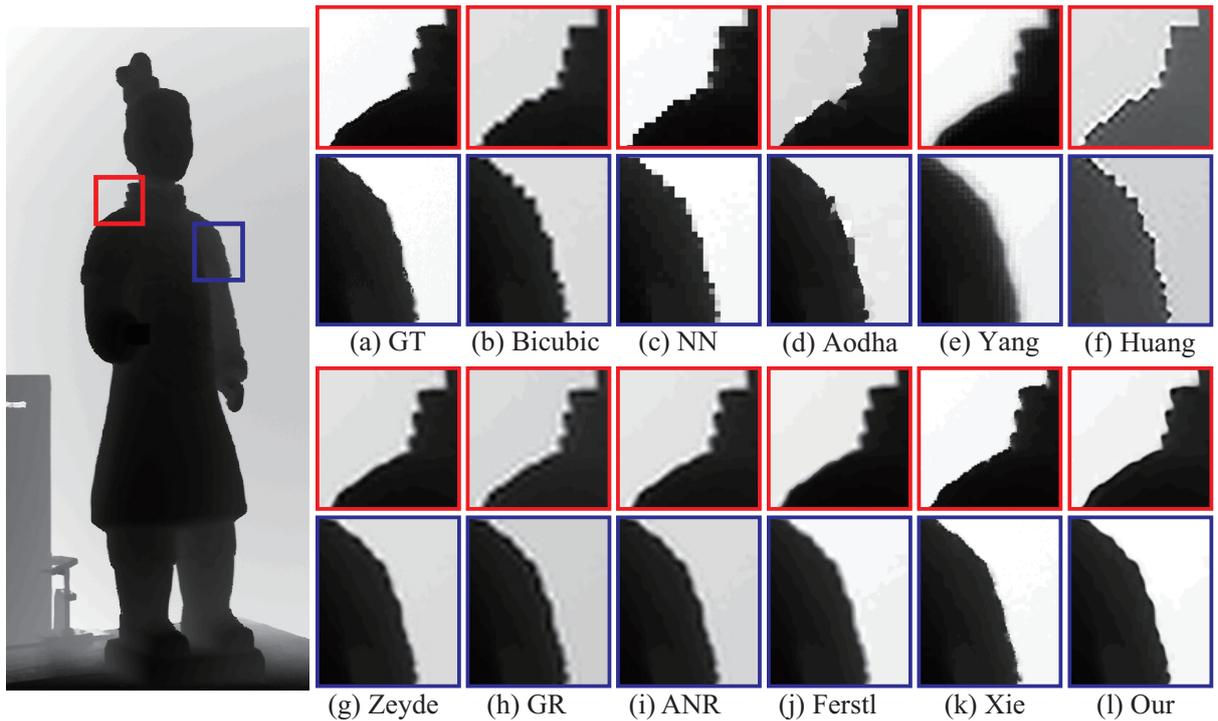

Fig. 9. Laser scan results (factor ×4). (a) Ground truth. (b) Bicubic. (c) Nearest Neighbor. (d) Aodha et al. [10]. (e) Yang et al. [25]. (f) Huang et al. [20]. (g) Zeyde et al. [26]. (h) GR [27]. (i) ANR [27]. (j) Ferstl et al. [12]. (k) Xie et al. [13]. (f) Our method. **Best viewed on screen.**

results.

To evaluate the train times, experiment has been done. The average quantitative results of 1st train, 2nd train and 3rd train on different data-sets in $RMSE$ for up-sampling factor of ×8 are shown in figure 5. We extract 4 images from Middlebury dataset, 5 images from Sintel dataset and 5 images from ICL dataset, then, the average $RMSE$ in each dataset is calculated. It is clearly to find that the results of 2nd train are much smaller than the 1st train's, while the results of the 3rd train are very similar with the 2nd train's, which means much high frequency depth information can be learned in the 1st and 2nd train, while less useful information can be learned in the 3rd train. As a result, the train times in our method is set as 2.

### B. Quantitative Results

In this section, we evaluate the performance of the different $SOTA$ methods on publicly available benchmarks. We show the $RMSE$, $MAE$ and $SSIM$ results on Middlebury dataset ($Cones$, $Teddy$, $Tsukuba$ and $Venus$), Sintel dataset ($Alley$, $Ambush$, $Bandage$, $Caves$ and $Market$) and 5 images extracted from ICL dataset for up-sampling factors of ×2, ×3 and ×4. Additionally, we show the results for the real-world laser scan dataset ($Scan21$, $Scan30$, $Scan42$) provided by Aodha et al. [10] for an up-sampling factor of ×4. Table. I, table. II, table. III show the comparison results of $RMSE$, $SSIM$ and $MAE$ between our method and the state-of-the-art on $Middlebury$ and $LaserScan$ data-sets for factors of ×2 and ×4 respectively. To further show the performance of our method, we also show the comparison results of $RMSE$ and $SSIM$ in table. IV and table. V on $Sintel$ and $ICL$ data-sets for factor of ×3. Numbers in bold indicate the best performance and those with underline indicate the second best performance. What can be clearly seen is that our proposed methods not only generate much smaller errors in $RMSE$, $MAE$, but also generate much better results in $SSIM$, which means that our method can well recover the HR depth maps well in both numerical value and structure.

### C. Qualitative Results

To better evaluate our method, we show different visual comparison results on different benchmarks with different up-scale factors, as shown in Fig. 6, 7, 8 and 9. Among these figures, Fig. 6 shows the results of the $Middlebury$ data (parts of $Cones$ and $Teddy$) with up-sampling factor of ×4. Fig. 9 shows the result of up-scaling the depth from a $Laser\ Scan$ with the zoomed cropped regions with the factor of ×4, and Fig. 7 shows the result of up-scaling the depth from $Sintel$ data-set ($Alley$ and $Caves$) for factor of ×4. Fig. 8 shows the result of up-scaling the depth from ICL data-set(part of office room 722) for factor of ×3. Note that image enhancement methods are employed in these depth maps in order to show the details more clearly. It is obvious to see that our method produces more visual appealing results than the previously reported approaches. Boundaries in the results generated by our method are sharper and smoother along the edge direction. Besides, our method can preserve the structure of the scene in regions with fine structures effectively.

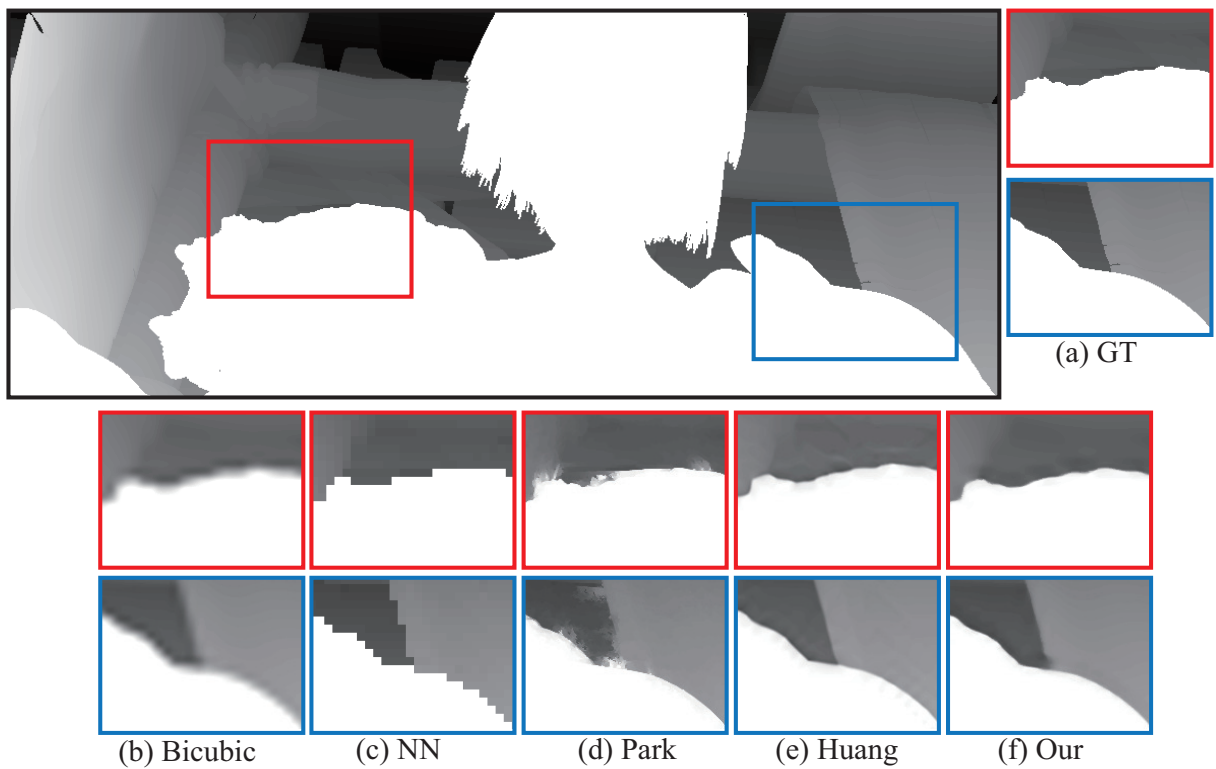

Fig. 10. Sintel result (factor ×8). (a) Ground Truth. (b) Bicubic. (c) Nearest neighbor. (d) Park et al. [14]. (e) Huang et al. [20]. (f) Our method. **Best viewed on screen.**

*D. Comparison on large upscale factor*

Many state-of-the-art methods fail to up-scale depth maps for large factors, such as ×8. Hence, to further show the effectiveness of our method in handling large up-scale factors, we show the quantitative and qualitative results for factor of ×8, which means infer $8 \times 8 = 64$ depth values from a single depth value. Table.VI shows the comparison results of $RMSE$ and $SSIM$ on $Middlebury$ data-set ($Cones$, $Teddy$, $Tsukuba$ and $Venus$) for factors of ×8, from which, we can see that our method generates better $RMSE$ and $SSIM$ scores. Compared with other methods, more boundary and structure information can be recovered by our method. Thus, our method outperforms other approaches even under large factors. Meanwhile, Fig. 10 shows the visual comparison results between our method and state-of-the-art methods. It is obviously to find that our method generates more sharper boundaries and the structure of the generated results preserved better than other methods for factor of ×8.

## V. Conclusions

In this work we propose a method for single depth image super-resolution, which can be divided into two steps. First, an end-to-end progressive deep convolutional network is used to generate high resolution depth images from low resolution depth images. To the best of our knowledge, we are the first one to use deep convolutional network to solve the problem of depth super-resolution. Second, to further improve the quality of the obtained high resolution depth images, depth statistical information and the responding high resolution color images are recognized as effective regularization to refine the learned depth images. As demonstrated in the experiment, we find that the depth map itself can even be employed to improve the quality of depth map. By combining the two steps, we are able to generate better depth images than state-of-the-art approaches. In a quantitative and qualitative evaluation using widespread data-sets we show that our method qualitatively outperforms existing methods.

## Acknowledgments

X. Song is supported by 863 Program of China (No. 2015AA016405) and China Scholarship Council. Y. Dai is funded in part by ARC Grants (DE140100180, LP100100588) and National Natural Science Foundation of China (61420106007). X. Qin is supported by 863 Program of China (No. 2015AA016405)

## References

[1] Shotton, J., Sharp, T., Kipman, A., Fitzgibbon, A., Finocchio, M., Blake, A., Cook, M., Moore, R.: Real-time human pose recognition in parts from single depth images. Communications of the ACM **56** (2013) 116–124
[2] Izadi, S., Kim, D., Hilliges, O., Molyneaux, D., Newcombe, R., Kohli, P., Shotton, J., Hodges, S., Freeman, D., Davison, A., et al.: Kinectfusion: real-time 3d reconstruction and interaction using a moving depth camera. In: Proceedings of the 24th annual ACM symposium on User interface software and technology, ACM (2011) 559–568
[3] Butler, D.J., Wulff, J., Stanley, G.B., Black, M.J.: A naturalistic open source movie for optical flow evaluation. In A. Fitzgibbon et al. (Eds.), ed.: European Conf. on Computer Vision (ECCV). Part IV, LNCS 7577, Springer-Verlag (2012) 611–625

| Method | $RMSE$ | | | | |
|---|---|---|---|---|---|
| | alley | ambush | bandage | cave | market |
| NN | 2.0304 | 9.3208 | 4.4105 | 8.7593 | 4.9647 |
| Bicubic | 1.6098 | 7.3625 | 3.4488 | 6.9150 | 4.0863 |
| Park et al. [14] | 1.8205 | 6.0661 | 3.5613 | 7.2288 | 4.6807 |
| Yang et al. [31] | 1.3415 | 6.2249 | 2.9892 | 5.8684 | 3.4367 |
| Ferstl et al. [16] | 1.7905 | 6.0874 | 3.5316 | 7.0437 | 3.8448 |
| Yang et al. [25] | 1.8846 | 7.1263 | 3.4441 | 6.8106 | 4.0149 |
| Zeyde et al. [26] | 1.3635 | 5.8015 | 2.9337 | 5.5711 | 3.3339 |
| GR [27] | 1.5227 | 6.8524 | 3.2246 | 6.4512 | 3.8334 |
| ANR [27] | 1.4385 | 6.2703 | 3.0565 | 5.9875 | 3.5555 |
| NE+LS | 1.4030 | 5.9957 | 2.9809 | 5.8207 | 3.4579 |
| NE+NNLS | 1.4315 | 6.2489 | 3.0448 | 6.0319 | 3.5313 |
| NE+LLE | 1.4744 | 6.3343 | 3.0692 | 6.0195 | 3.5682 |
| Aodha et al. [10] | 2.4248 | 7.7369 | 3.7988 | 7.4757 | 6.2797 |
| Huang et al. [20] | 2.0955 | 10.0955 | 5.4169 | 9.5219 | 5.4694 |
| Xie et al. [13] | 1.7346 | 6.5696 | 3.4876 | 6.8372 | 3.7909 |
| Our | **1.0128** | **2.7181** | **2.0646** | **2.9314** | **1.8106** |
| Method | $SSIM$ | | | | |
| | alley | ambush | bandage | cave | market |
| NN | 0.9775 | 0.9659 | 0.9773 | 0.9495 | 0.9720 |
| Bicubic | 0.9867 | 0.9749 | 0.9844 | 0.9644 | 0.9801 |
| Park et al. [14] | 0.9814 | 0.9791 | 0.9831 | 0.9588 | 0.9784 |
| Yang et al. [31] | 0.9870 | 0.9769 | 0.9852 | 0.9671 | 0.9815 |
| Ferstl et al. [16] | 0.9763 | 0.9643 | 0.9760 | 0.9463 | 0.9671 |
| Yang et al. [25] | 0.9851 | 0.9752 | 0.9873 | 0.9691 | 0.9816 |
| Zeyde et al. [26] | 0.9907 | 0.9843 | 0.9896 | 0.9779 | 0.9874 |
| GR [27] | 0.9875 | 0.9711 | 0.9850 | 0.9621 | 0.9794 |
| ANR [27] | 0.9895 | 0.9804 | 0.9881 | 0.9728 | 0.9847 |
| NE+LS | 0.9901 | 0.9839 | 0.9893 | 0.9770 | 0.9866 |
| NE+NNLS | 0.9891 | 0.9806 | 0.9886 | 0.9733 | 0.9849 |
| NE+LLE | 0.9865 | 0.9793 | 0.9860 | 0.9724 | 0.9830 |
| Aodha et al. [10] | 0.9732 | 0.9635 | 0.9763 | 0.9432 | 0.9694 |
| Huang et al. [20] | 0.9771 | 0.9485 | 0.9677 | 0.9353 | 0.9663 |
| Xie et al. [13] | 0.9810 | 0.9883 | 0.9827 | 0.9786 | 0.9869 |
| Our | **0.9916** | **0.9936** | **0.9931** | **0.9904** | **0.9936** |

TABLE IV
Quantitative evaluation. The $RMSE$ and $SSIM$ are calculated for different SOTA methods on the Sintel data-set for factors of ×3. The best result of all single image methods for each dataset and up-scaling factor is highlighted and the second best is underlined.

| Method | $RMSE$ | | | | |
|---|---|---|---|---|---|
| | 308 | 344 | 531 | 54 | 722 |
| NN | 1.4935 | 1.0300 | 1.3506 | 0.8322 | 1.1613 |
| Bicubic | 1.0261 | 0.6751 | 0.8890 | 0.5696 | 0.8089 |
| Park et al. [14] | 1.2735 | 0.8728 | 1.0174 | 0.7608 | 0.9636 |
| Yang et al. [31] | 0.9584 | 0.6355 | 0.8490 | 0.5586 | 0.7566 |
| Ferstl et al. [16] | 1.2533 | 0.7623 | 0.9971 | 0.6894 | 0.9537 |
| Yang et al. [25] | 0.9351 | 0.6760 | 0.8692 | 0.6092 | 0.8384 |
| Zeyde et al. [26] | 0.8124 | 0.4982 | 0.6756 | 0.4592 | 0.6659 |
| GR [27] | 0.9326 | 0.6409 | 0.8369 | 0.5377 | 0.7567 |
| ANR [27] | 0.8567 | 0.5838 | 0.7573 | 0.4916 | 0.7091 |
| NE+LS | 0.8278 | 0.5317 | 0.7056 | 0.4733 | 0.6858 |
| NE+NNLS | 0.8788 | 0.5726 | 0.7693 | 0.4942 | 0.7175 |
| NE+LLE | 0.9982 | 0.7780 | 0.9548 | 0.7260 | 0.8430 |
| Aodha et al. [10] | 1.6565 | 1.2548 | 1.5698 | 0.9113 | 1.3240 |
| Huang et al. [20] | 1.5419 | 1.0231 | 1.3788 | 0.8770 | 1.2410 |
| Xie et al. [13] | 1.1549 | 0.8551 | 1.0381 | 0.5902 | 0.9294 |
| Our | **0.7127** | **0.4038** | **0.4910** | **0.4320** | **0.6318** |
| Method | $SSIM$ | | | | |
| | 308 | 344 | 531 | 54 | 722 |
| NN | 0.9918 | 0.9950 | 0.9921 | 0.9950 | 0.9904 |
| Bicubic | 0.9949 | 0.9973 | 0.9957 | 0.9973 | 0.9943 |
| Park et al. [14] | 0.9955 | 0.9971 | 0.9965 | 0.9972 | 0.9954 |
| Yang et al. [31] | 0.9952 | 0.9972 | 0.9967 | 0.9971 | 0.9949 |
| Ferstl et al. [16] | 0.9934 | 0.9965 | 0.9948 | 0.9964 | 0.9934 |
| Yang et al. [25] | 0.9949 | 0.9966 | 0.9947 | 0.9965 | 0.9938 |
| Zeyde et al. [26] | 0.9968 | 0.9984 | 0.9974 | 0.9982 | 0.9962 |
| GR [27] | 0.9955 | 0.9974 | 0.9959 | 0.9974 | 0.9948 |
| ANR [27] | 0.9963 | 0.9979 | 0.9967 | 0.9979 | 0.9955 |
| NE+LS | 0.9966 | 0.9982 | 0.9972 | 0.9980 | 0.9959 |
| NE+NNLS | 0.9961 | 0.9979 | 0.9965 | 0.9978 | 0.9955 |
| NE+LLE | 0.9925 | 0.9939 | 0.9916 | 0.9938 | 0.9925 |
| Aodha et al. [10] | 0.9905 | 0.9941 | 0.9913 | 0.9948 | 0.9890 |
| Huang et al. [20] | 0.9913 | 0.9950 | 0.9917 | 0.9947 | 0.9898 |
| Xie et al. [13] | 0.9784 | 0.9846 | 0.9869 | 0.9819 | 0.9725 |
| Our | **0.9975** | **0.9987** | **0.9983** | **0.9983** | **0.9966** |

TABLE V
Quantitative evaluation. The $RMSE$ and $SSIM$ are calculated for different SOTA methods on the ICL dataset for factors of ×3. The best result of all single image methods for each data-set and up-scaling factor is highlighted and the second best is underlined.

[4] Dong, C., Loy, C.C., He, K., Tang, X.: Learning a deep convolutional network for image super-resolution. In: Computer Vision–ECCV 2014. Springer (2014) 184–199
[5] Kim, J., Lee, J.K., Lee, K.M.: Accurate image super-resolution using very deep convolutional networks. arXiv preprint arXiv:1511.04587 (2015)
[6] Kim, J., Lee, J.K., Lee, K.M.: Deeply-recursive convolutional network for image super-resolution. arXiv preprint arXiv:1511.04491 (2015)
[7] Schuon, S., Theobalt, C., Davis, J., Thrun, S.: Lidarboost: Depth superresolution for tof 3d shape scanning. In: Computer Vision and Pattern Recognition, 2009. CVPR 2009. IEEE Conference on, IEEE (2009) 343–350
[8] Rajagopalan, A., Bhavsar, A., Wallhoff, F., Rigoll, G.: Resolution enhancement of pmd range maps. In: Pattern Recognition. Springer (2008) 304–313
[9] Freeman, W.T., Jones, T.R., Pasztor, E.C.: Example-based super-resolution. Computer Graphics and Applications, IEEE 22 (2002) 56–65
[10] Mac Aodha, O., Campbell, N.D., Nair, A., Brostow, G.J.: Patch based synthesis for single depth image super-resolution. In: Proc. Eur. Conf. Comp. Vis. Springer (2012) 71–84
[11] Hornácek, M., Rhemann, C., Gelautz, M., Rother, C.: Depth super resolution by rigid body self-similarity in 3d. In: Proceedings of the IEEE Conference on Computer Vision and Pattern Recognition. (2013) 1123–1130
[12] Ferstl, D., Ruther, M., Bischof, H.: Variational depth superresolution using example-based edge representations. In: Proceedings of the IEEE International Conference on Computer Vision. (2015) 513–521
[13] Xie, J., Feris, R.S., Sun, M.T.: Edge-guided single depth image super resolution. Image Processing, IEEE Transactions on 25 (2016) 428–438
[14] Park, J., Kim, H., Tai, Y.W., Brown, M.S., Kweon, I.: High quality depth map upsampling for 3d-tof cameras. In: Computer Vision (ICCV), 2011 IEEE International Conference on, IEEE (2011) 1623–1630
[15] Yang, J., Ye, X., Li, K., Hou, C.: Depth recovery using an adaptive color-guided auto-regressive model. In: Computer Vision–ECCV 2012. Springer (2012) 158–171

| Method | $RMSE$ | | | |
|---|---|---|---|---|
| | Cones | Teddy | Tsukuba | Venus |
| NN | 7.5937 | 6.2416 | 18.4786 | 4.4645 |
| Bicubic | 5.3000 | 4.2423 | 13.3220 | 2.8948 |
| Park et al. [14] | 8.0078 | 6.3264 | 17.6625 | 3.4086 |
| Yang et al. [31] | 5.1390 | 4.0660 | 13.1748 | 2.7559 |
| Huang et al. [20] | 6.1629 | 6.6235 | **10.6618** | 4.1399 |
| Our | **4.5887** | **2.88497** | 11.6231 | **1.7082** |
| Method | $SSIM$ | | | |
| | Cones | Teddy | Tsukuba | Venus |
| NN | 0.8996 | 0.9199 | 0.8387 | 0.9634 |
| Bicubic | 0.9314 | 0.9442 | 0.8564 | 0.9757 |
| Park et al. [14] | 0.9231 | 0.9426 | 0.8409 | 0.9792 |
| Yang et al. [31] | 0.9361 | 0.9482 | 0.8624 | 0.9768 |
| Huang et al. [20] | 0.9280 | 0.9254 | 0.9027 | 0.9712 |
| Our | **0.9510** | **0.9679** | **0.9051** | **0.9903** |

TABLE VI
Quantitative evaluation. The $RMSE$ and $SSIM$ are calculated for different SOTA methods for the Middlebury data-set for factors of and ×8. The best result of all single image methods for each data-set and up-scaling factor is highlighted and the second best is underlined.

[16] Ferstl, D., Reinbacher, C., Ranftl, R., Rüther, M., Bischof, H.: Image guided depth upsampling using anisotropic total generalized variation. In: Proceedings of the IEEE International Conference on Computer Vision. (2013) 993–1000
[17] Matsuo, K., Aoki, Y.: Depth image enhancement using local tangent plane approximations. In: Computer Vision and Pattern Recognition (CVPR), 2015 IEEE Conference on, IEEE (2015) 3574–3583
[18] Lu, J., Forsyth, D.: Sparse depth super resolution. In: Computer Vision and Pattern Recognition (CVPR), 2015 IEEE Conference on, IEEE (2015) 2245–2253
[19] Glasner, D., Bagon, S., Irani, M.: Super-resolution from a single image. In: Computer Vision, 2009 IEEE 12th International Conference on, IEEE (2009) 349–356

[20] Huang, J.B., Singh, A., Ahuja, N.: Single image super-resolution from transformed self-exemplars. In: Computer Vision and Pattern Recognition (CVPR), 2015 IEEE Conference on, IEEE (2015) 5197–5206

[21] Yang, J., Wright, J., Huang, T.S., Ma, Y.: Image super-resolution via sparse representation. Image Processing, IEEE Transactions on **19** (2010) 2861–2873

[22] Wang, S., Zhang, L., Liang, Y., Pan, Q.: Semi-coupled dictionary learning with applications to image super-resolution and photo-sketch synthesis. In: Computer Vision and Pattern Recognition (CVPR), 2012 IEEE Conference on, IEEE (2012) 2216–2223

[23] Kim, K.I., Kwon, Y.: Single-image super-resolution using sparse regression and natural image prior. Pattern Analysis and Machine Intelligence, IEEE Transactions on **32** (2010) 1127–1133

[24] Yang, M.C., Wang, Y.C.F.: A self-learning approach to single image super-resolution. Multimedia, IEEE Transactions on **15** (2013) 498–508

[25] Yang, J., Wright, J., Huang, T.S., Ma, Y.: Image super-resolution via sparse representation. **19** (2010) 2861–2873

[26] Zeyde, R., Elad, M., Protter, M.: On single image scale-up using sparse-representations. In: Curves and Surfaces. Springer (2010) 711–730

[27] Timofte, R., Smet, V., Gool, L.: Anchored neighborhood regression for fast example-based super-resolution. In: Proc. IEEE Int. Conf. Comp. Vis. (2013) 1920–1927

[28] Levin, A., Lischinski, D., Weiss, Y.: Colorization using optimization. In: ACM Transactions on Graphics (TOG). Volume 23., ACM (2004) 689–694

[29] Silberman, N., Hoiem, D., Kohli, P., Fergus, R.: Indoor segmentation and support inference from rgbd images. In: Computer Vision–ECCV 2012. Springer (2012) 746–760

[30] Diebel, J., Thrun, S.: An application of markov random fields to range sensing. In: NIPS. Volume 5. (2005) 291–298

[31] Yang, J., Ye, X., Li, K., Hou, C.: Depth recovery using an adaptive color-guided auto-regressive model. In: Computer Vision–ECCV 2012. Springer (2012) 158–171

[32] Chartrand, R., Yin, W.: Iteratively reweighted algorithms for compressive sensing. In: Acoustics, speech and signal processing, 2008. ICASSP 2008. IEEE international conference on, IEEE (2008) 3869–3872

[33] Ajanthan, T., Hartley, R., Salzmann, M., Li, H.: Iteratively reweighted graph cut for multi-label mrfs with non-convex priors. In: Proceedings of the IEEE Conference on Computer Vision and Pattern Recognition. (2015) 5144–5152

[34] Scharstein, D., Szeliski, R.: A taxonomy and evaluation of dense two-frame stereo correspondence algorithms. International journal of computer vision **47** (2002) 7–42

[35] Scharstein, D., Szeliski, R.: High-accuracy stereo depth maps using structured light. In: Computer Vision and Pattern Recognition, 2003. Proceedings. 2003 IEEE Computer Society Conference on. Volume 1., IEEE (2003) I–195

[36] Scharstein, D., Pal, C.: Learning conditional random fields for stereo. In: Computer Vision and Pattern Recognition, 2007. CVPR'07. IEEE Conference on, IEEE (2007) 1–8

[37] Hirschmüller, H., Scharstein, D.: Evaluation of cost functions for stereo matching. In: Computer Vision and Pattern Recognition, 2007. CVPR'07. IEEE Conference on, IEEE (2007) 1–8

[38] Handa, A., Whelan, T., McDonald, J., Davison, A.: A benchmark for RGB-D visual odometry, 3D reconstruction and SLAM. In: IEEE Intl. Conf. on Robotics and Automation, ICRA, Hong Kong, China (2014)

[39] Bevilacqua, M., Roumy, A., Guillemot, C., Alberi-Morel, M.L.: Low-complexity single-image super-resolution based on nonnegative neighbor embedding. (2012)